%% file: root.tex
\documentclass[letterpaper, 10 pt, conference]{ieeeconf}

\IEEEoverridecommandlockouts                           
\usepackage{times}
\usepackage{graphics} 
\usepackage{mathtools}
\usepackage{graphicx}
\usepackage{tabularx}
\usepackage{caption}
\usepackage{wrapfig}
\usepackage{romannum}
\usepackage{amsmath,amssymb}
\usepackage{url}
\usepackage{bm}
\usepackage[table,xcdraw]{xcolor}
\usepackage{hyperref}
\usepackage{tabularray}
\usepackage{cite}

\usepackage{multirow}
\usepackage{booktabs}
\usepackage{pifont}
\usepackage{arydshln}
\usepackage{tabularx}

\usepackage[ruled,vlined]{algorithm2e}
\usepackage[noend]{algpseudocode}
\makeatletter
\let\OldStatex\Statex
\renewcommand{\Statex}[1][3]{%
  \setlength\@tempdima{\algorithmicindent}%
  \OldStatex\hskip\dimexpr#1\@tempdima\relax}
\renewcommand{\ALG@beginalgorithmic}{\normalsize}

\newcommand{\cmark}{\ding{51}}%
\newcommand{\xmark}{\ding{55}}%

\DeclareCaptionFont{mysize}{\fontsize{8}{9.6}\selectfont}
\title{\LARGE \bf
DA-RAW: Domain Adaptive Object Detection for Real-World \\ Adverse Weather Conditions
}

\author{Minsik Jeon$^{*}$, Junwon Seo$^{*}$, Jihong Min%
\thanks{This work was supported by the Agency For Defense Development Grant funded by the Korean Government in 2023.}
\thanks{Minsik Jeon, Junwon Seo, and Jihong Min are with the Agency for Defense Development, Republic of Korea
        {\tt\footnotesize \{mikejeon001123, junwon.vision, happymin77\}@gmail.com}}%
\thanks{$^{*}$These authors contributed equally to this work.}
\thanks{Our project website can be found at \href{http://bit.ly/3yccTRa}{\tt\footnotesize http://bit.ly/3yccTRa}}%
}
\begin{document}

\maketitle

\begin{abstract}
Despite the success of deep learning-based object detection methods in recent years, it is still challenging to make the object detector reliable in adverse weather conditions such as rain and snow. For the robust performance of object detectors, unsupervised domain adaptation has been utilized to adapt the detection network trained on clear weather images to adverse weather images. While previous methods do not explicitly address weather corruption during adaptation, the domain gap between clear and adverse weather can be decomposed into two factors with distinct characteristics: a style gap and a weather gap. In this paper, we present an unsupervised domain adaptation framework for object detection that can more effectively adapt to real-world environments with adverse weather conditions by addressing these two gaps separately. Our method resolves the style gap by concentrating on style-related information of high-level features using an attention module. Using self-supervised contrastive learning, our framework then reduces the weather gap and acquires instance features that are robust to weather corruption. Extensive experiments demonstrate that our method outperforms other methods for object detection in adverse weather conditions.
\end{abstract}

\section{INTRODUCTION}
\input{_I.Introduction/1.1.introduction}

\section{RELATED WORKS}
\subsection{Object Detection in Adverse Weather Condition}
\input{_II.Related_Work/2.1.Object_Detection}

\subsection{Unsupervised Domain Adaptation for Object Detection}
\input{_II.Related_Work/2.2.Domain_Adaptation}
\section{METHODS}

\subsection{Preliminaries}
\input{_III.Method/3.1.Detection_Network}

\subsection{Image-level Style Alignment}
\input{_III.Method/3.2.Image_Align}

\subsection{Instance-level Weather Alignment}
\input{_III.Method/3.3.Instance_Contrast}

\section{EXPERIMENTS}
\input{_IV.Experiments/4.0.Overview}

\subsection{Datasets}
\input{_IV.Experiments/4.1.Datasets}

\subsection{Experimental Setup}
\input{_IV.Experiments/4.2.Experimental_setup}

\subsection{Experimental Results}
\input{_IV.Experiments/4.3.Experimental_Results}

\input{_IV.Experiments/4.4.Analysis}

\section{CONCLUSION}
\input{_V.Conclusion/5.1.Conclusion}

\addtolength{\textheight}{0cm}   

\bibliographystyle{IEEEtran}
\bibliography{mybib.bib}

\end{document}

%% file: _I.Introduction/1.1.Introduction.tex
Object detection plays a crucial role in enabling machines, such as autonomous vehicles and surveillance systems, to perceive and comprehend their surrounding environment. While deep learning has significantly improved object detection capabilities, ensuring the accuracy of these systems under adverse weather conditions like rain and snow remains an ongoing challenge. To ensure the detector's dependability, it is necessary to develop a learning method that can adapt object detectors to adverse weather conditions.

Due to the laborious process of obtaining labeled data for real-world adverse weather conditions, various methods have utilized synthetic datasets to improve detection performance. By generating synthetic weather effects on clear weather images without degradation, fully annotated images of adverse weather are obtained. These images are utilized to train the robust model in a supervised manner~\cite{halder2019physics,tremblay2021rain,volk2019towards}, or the removal network can be trained to restore a clear image from adverse weather images~\cite{qian2018attentive,wang2021rain,ye2021closing,ye2022unsupervised}. However, prior knowledge of weather conditions cannot effectively capture the intricate characteristics of adverse weather conditions in the real world, which have diverse and complex effects on images. Therefore, relying on synthetic datasets does not significantly enhance the model's performance when applied to real-world environments. Recent works also suggest that separately trained removal networks do not help downstream tasks~\cite{pei2018does,li2019single}, implying the need for methods that improve the performance of downstream tasks in adverse weather conditions.

\definecolor{StyleRed}{HTML}{E83F3F}
\definecolor{WeatherBlue}{HTML}{0091FF}

\begin{figure}[t]
\centering
\includegraphics[width=1.0\linewidth]{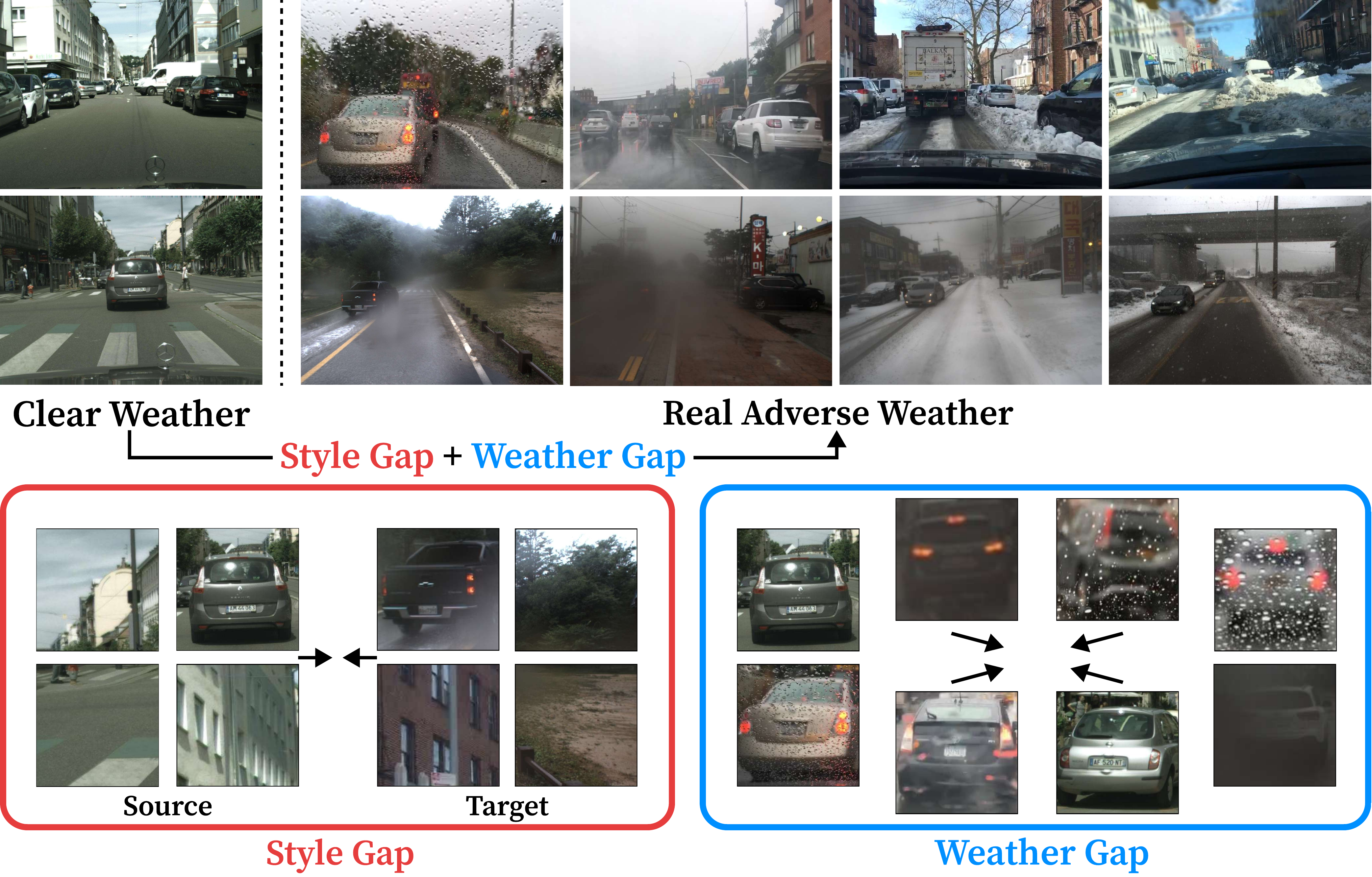}
\caption{
We propose a novel unsupervised domain adaptation method capable of adapting an object detector from clear weather to real-world adverse weather conditions with a significant domain gap. This gap can be divided into two distinct factors: the \textcolor{StyleRed}{\textit{Style Gap}} and the \textcolor{WeatherBlue}{\textit{Weather Gap}}. The \textcolor{StyleRed}{\textit{Style Gap}} stems from environmental changes such as the image's background or color, whereas the \textcolor{WeatherBlue}{\textit{Weather Gap}} is caused by weather corruptions like rain stains, which introduce random and localized image degradation. Due to the distinct characteristics of the two gaps, we employ separate modules to address each of them independently.
}
\label{fig:concept}
\end{figure}

Recent studies have focused on Unsupervised Domain Adaptation~(UDA) to enhance the robustness of object detectors in adverse weather conditions~\cite{chen2018domain,saito2019strong,sindagi2020prior,xu2020cross,deng2021unbiased,li2022cross,cao2023contrastive,li2023domain}. These methods adapt the model trained in the source domain of clear weather to the target domain of adverse weather by considering adverse weather as a factor contributing to the domain gap~\cite{hnewa2020object,oza2023unsupervised}. Without requiring the ground truth labels of target domain images, most UDA methods align the feature distributions of the two domains globally in an adversarial manner~\cite{chen2018domain,saito2019strong,sindagi2020prior,xu2020cross,chen2021scale}.

While most UDA methods regard the domain gap between clear and adverse weather data similarly to conventional domain adaptation settings, the gap can be broken down into two distinct factors: the \textit{style gap} and the \textit{weather gap}~\cite{ma2022both}, as shown in Fig~\ref{fig:concept}. Style gaps are caused by variations in the operating environment (e.g., background, color, texture), whereas weather gaps result from weather-induced corruption (e.g., rain stains, snowflakes). 
Unlike style gaps, which are caused by global and semantic factors, weather corruption produces arbitrary and localized image degradation that is hard to characterize using prior knowledge~\cite{li2019single}.
Existing UDA methods consider weather corruption as part of the image's style and align the source and target distributions globally. As some features are arbitrarily and severely distorted by weather corruption, these methods frequently lead to suboptimal alignment under adverse weather conditions. Consequently, they are only effective on synthetic datasets with minor domain gaps~\cite{chen2018domain}, whereas their performance degrades when applied to real-world datasets with a large style gap and complex weather corruption~\cite{ma2022both}. Separately addressing the two aspects of the domain gap improves domain alignment and enables robust object detection in real-world adverse weather conditions.

In this paper, we propose an unsupervised domain adaptation method to enhance the robustness of object detection in real-world adverse weather conditions. Specifically, we resolve the style and weather gaps separately to achieve optimal feature alignment. To bridge the style gap, our method aligns high-level style-related features using an attention module. Moreover, self-supervised contrastive learning is employed to resolve the weather gap. Based on the assumption that each instance consists of an object and random weather corruption, our model encourages the similarity between instance features within the same class, resulting in a robust representation against corruption. To demonstrate the efficacy of our method in a variety of real-world scenarios, we collect actual driving data in a wide range of environments and weather conditions. Through extensive experiments, we demonstrate that our method effectively adapts to various real-world datasets.

%% file: _II.Related_Work/2.1.Object_Detection.tex
For the reliable perception of environments, numerous methods attempt to train robust object detectors under adverse weather~\cite{hnewa2020object,mirza2021robustness,wu2022contrastive,rothmeier2023had}.
The most intuitive approach is to utilize annotated datasets of adverse weather conditions~\cite{li2019single,hassaballah2020vehicle,yu2020bdd100k}. Due to the difficulty of acquiring labeled data for real-world adverse weather, synthetic weather effects are generated on labeled clear images using prior knowledge of image formulation under adverse weather condition~\cite{halder2019physics,volk2019towards,tremblay2021rain}. The object detector is then trained in a supervised manner using this synthetic dataset. Other methods train removal networks to restore clear images from adverse weather images using paired data of clear and synthetic weather images with the same background~\cite{qian2018attentive,wang2019spatial,zamir2021multi}, or using unpaired data~\cite{wei2021deraincyclegan,chen2022unpaired,ye2022unsupervised}. To acquire more realistic synthetic data, some methods jointly train a synthetic data generation model and its removal network~\cite{wang2021rain,ye2021closing}.

In real-world environments, however, the efficacy of methods that utilize synthetic data decreases due to the complexity and diversity of real-world weather corruptions~\cite{li2019single}. In addition, removal networks are computationally intensive to be attached to the front of the detection network, and they are trained independently to downstream tasks, which provides insufficient performance improvement for these tasks on real-world images~\cite{pei2018does,li2019single}. While some methods have attempted to jointly train the removal network and downstream tasks~\cite{lee2022perception, wang2022end,liu2022image,kalwar2023gdip}, they still rely on synthetic data or impose a computational burden.

%% file: _II.Related_Work/2.2.Domain_Adaptation.tex
Unsupervised domain adaptation can be used to directly adapt a detector trained on the source domain of clear weather to the target domain of adverse weather~\cite{deng2021unbiased,eskandar2022unsupervised,li2022cross,zhang2022multiple,cao2023contrastive,oza2023unsupervised}. Most UDA methods jointly train a domain classifier and a detector so that the classifier distinguishes between the source and target features, while the detector is optimized to confuse the classifier and align the feature distributions globally~\cite{ganin2016domain,chen2018domain}. Alignment can be performed on image-level features from various backbones, such as ResNet~\cite{saito2019strong} or Feature Pyramid Network~(FPN)~\cite{chen2021scale}. In addition, aligning instance-level features extracted from Region-of-Interest~(RoI) can improve domain alignment~\cite{xu2020cross,rezaeianaran2021seeking, vs2021mega}.
Diverging from conventional UDA approaches, some methods employ mathematical formulations of adverse weather conditions to enhance the alignments of features~\cite{sindagi2020prior,li2023domain}. However, these methods consider the weather corruption as a part of an image's style and do not distinguish between the style and the weather gap~\cite{choi2021robustnet,lee2022fifo}, despite their distinct characteristics. This oversight results in suboptimal adaptation performance in real adverse weather conditions with both significant style and weather gaps~\cite{ma2022both}.

%% file: _III.Method/3.1.Detection_Network.tex
\begin{figure*}[t]
\begin{center}
\includegraphics[width=1.0\linewidth]{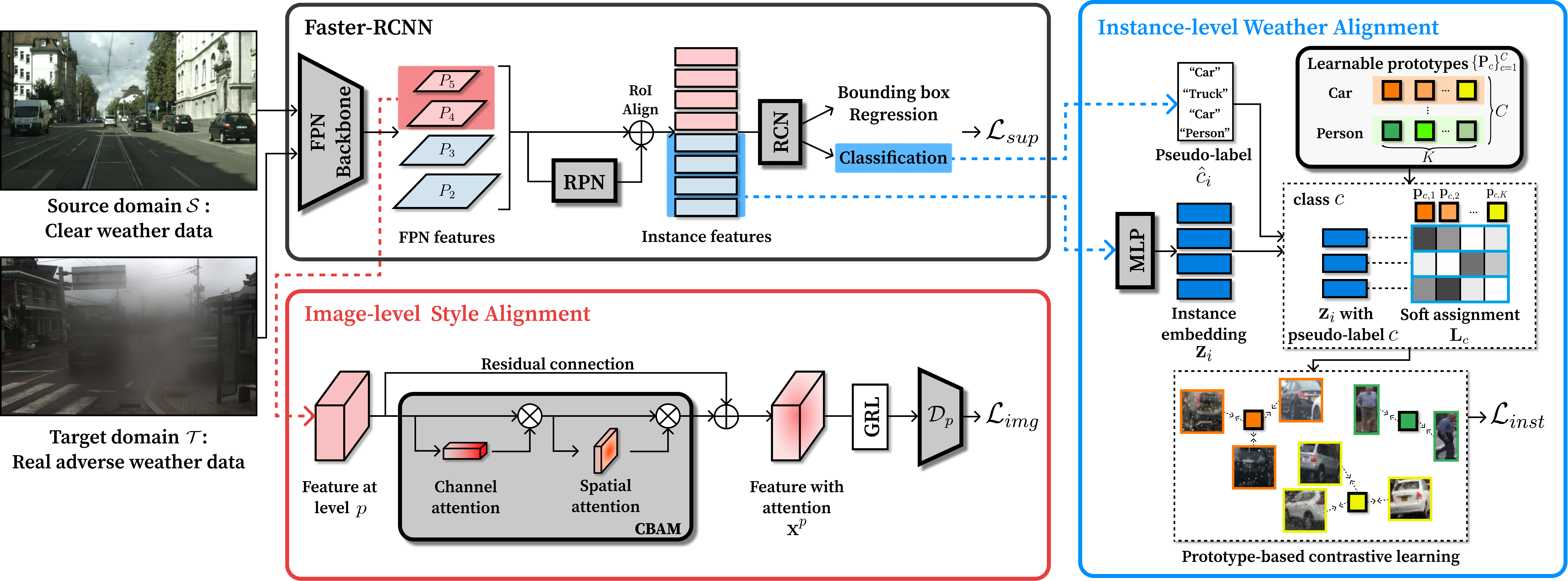}
\end{center}
\caption{Overall pipeline of the proposed method. Faster R-CNN with an FPN backbone is adopted for a detection network. \textcolor{StyleRed}{\textit{Image-level style alignment}} reduces the style gap by aligning the FPN's high-level features. During alignment, they focus on style-related features by incorporating CBAM and highlighting important spatial and channel details. \textcolor{WeatherBlue}{\textit{Instance-level weather alignment}} uses instance embedding and its corresponding pseudo-label from RCN to establish a soft assignment for each feature to learnable class prototypes. Using multi-prototype-based contrastive learning, it resolves the weather gap and constructs a weather-resistant feature representation by increasing the similarity between an instance embedding and its assigned prototypes.
}
\label{fig:model}
\end{figure*}

Given labeled images of clear weather conditions from the source domain $\mathcal{S}$ and unlabeled images of adverse weather conditions from the target domain $\mathcal{T}$, each minibatch consists of the same number of source and target data. Note that the source and target data are distinct in terms of both weather conditions and the surrounding environment.

We utilize the Faster R-CNN~\cite{ren2015faster} pipeline with an FPN~\cite{lin2017feature} backbone during training. The FPN backbone employs pyramid architecture to generate multi-scale feature maps $(P_2, P_3, P_4, P_5)$ from an image, allowing the efficient detection of objects of varying scales. The Region Proposal Network~(RPN) proposes RoI on these features and extracts instance features from each RoI, following the Region Classification Network~(RCN) which makes final class and bounding box predictions. The supervised loss $\mathcal{L}_{\text{sup}}$ obtained from RPN and RCN is applied only to the source data~\cite{chen2018domain}.

The overall architecture of our method is depicted in Fig~\ref{fig:model}. We aim to train a robust object detector that performs well in real-world environments with adverse weather. Based on the architecture of the FPN-based Faster R-CNN framework, we propose two components for domain adaptation to handle both style and weather gaps. First, an image-level style alignment is used to reduce the style gap through adversarial training. An instance-level weather alignment is then utilized to reduce the weather gap and learn the corruption-invariant features. The entire model is trained simultaneously in an end-to-end manner.

%% file: _III.Method/3.2.Image_Align.tex
The distinct styles of the source and target domains' environments result in different feature distributions. This style gap between domains is resolved through the alignment of image-level features. Similar to~\cite{chen2021scale}, a domain classifier is attached to each layer of the FPN backbone to distinguish between domains. The backbone is then trained to generate the domain-invariant feature by confusing the domain classifier through adversarial training. These objectives are accomplished in a single backpropagation step by a Gradient Reversal Layer~(GRL) with a weight coefficient $\lambda$.

We intend to perform image-level feature adaptation by focusing solely on the style properties of images. However, some features are severely degraded due to weather corruption, making it difficult to concentrate on the style differences. To enable the network to emphasize style-related features during alignment, the Convolutional Block Attention Module (CBAM)~\cite{woo2018cbam} is employed to emphasize features essential for domain alignment. CBAM is attached to each feature map and applies channel and spatial attention modules to acquire refined features $\mathbf{x}^p$ at feature level $p$. The attended feature is then fed into the discriminator $\mathcal{D}_p$ that predicts the domain of a feature, leading it to align features through the GRL by concentrating on essential information.

Since low-level features with fine-grained details are more susceptible to weather corruption, only high-level features are used for alignment. Therefore, image-level style alignment is performed on the $P_4$ and $P_5$ layers of the FPN backbone. The loss for image-level alignment, $\mathcal{L}_\text{img}$, is given by the following equation:
\begin{equation}
\begin{aligned}\label{img_align}
    \mathcal{L}_{\text{img}} = -\sum_{p} \sum_{\mathbf{x}_{i}^p}
            & \left[ y_i \log \mathcal{D}_p \left( \mathbf{x}_{i}^p \right) \right. \\ 
            & \left. + \left(1-y_i\right) \log \left(1-\mathcal{D}_p\left(\mathbf{x}_{i}^p\right)\right) \right],
\end{aligned}
\end{equation}
where $p \in \{P_4, P_5\}$ represents feature level, $\mathbf{x}_{i}^p$ represents each feature of level $p$ at location $i$ after CBAM layers, and $y_i \in \{0, 1\}$ indicates the domain label of each feature at location $i$.

%% file: _III.Method/3.3.Instance_Contrast.tex
The weather corruption has a local effect on the image and substantially degrades the instance features that are essential for object detection. We aim to obtain corruption-invariant features and reduce the weather gap through prototype-based contrastive learning. In particular, we assume that an instance feature of the target domain image is composed of an object and an arbitrary pattern of weather corruption. Then, the similarity between instance features of object proposals within the same category is encouraged, resulting in instance features invariant to weather corruption.

From the source and target domain training images, instance features are obtained, and each of them is pseudo-labeled as class $\hat{c_i}$ using the classwise score of each instance provided by the RCN. To be utilized for contrastive learning, instance features are forwarded to an MLP head to generate instance embeddings $\mathbf{Z} = \{\mathbf{z}_i\}_{i=1}^{N}$ with dimension $D$, where $N$ is the total number of instances. Note that MLPs exist independently for each level of features without sharing weights, and only instance features with scores over a threshold $\delta$ from the low-level image feature are utilized to align fine-grained features.

To maximize the similarity between instance embeddings with the same pseudo-labels, prototype-anchored metric learning is used to design the contrastive loss~\cite{zhou2022rethinking}. Using learnable prototypes as representatives of each class, each instance embedding is assigned to prototypes, and a network is learned to increase their similarity. $K$ learnable prototypes are used for each class $c$ as $\mathbf{P}_c \in \mathbb{R}^{D \times K}$ to account for the intra-class variation of instance features, and each prototype $\mathbf{p}_{c,k} \in \mathbb{R}^D$ serves as the $k^{th}$ cluster center of a class $c$. Also, to further boost the performance of contrastive learning, instance embeddings with the background class also adopt the same number of learnable prototypes, which can be used as negative samples for other instances.

Each instance with a pseudo-label $c$ is assigned to prototypes of the same class by computing the soft assignment matrix for class $c$, $\mathbf{L}_c \in \mathbb{R}_{+}^{K \times n_c}$. The soft assignment matrix satisfies the condition that the sum of soft assignment probabilities for each instance is one, \textit{i.e.}, $\mathbf{L}_c^{\top} \cdot \mathbf{1}^K = \mathbf{1}^{n_c}$, where $\mathbf{1}^K$ and $\mathbf{1}^{n_c}$ denotes the vector of all ones with dimensions $K$ and $n_c$, respectively. The assignment matrix can be obtained by maximizing the similarity between instance embeddings and the class prototypes, $\mathbf{Q}_c=\mathbf{P}_c^{\top}\mathbf{Z}_c\in\mathbb{R}^{K \times n_c}$, where $\mathbf{Z}_c\in\mathbb{R}^{D \times n_c}$ and $n_c$ represent instance embeddings and the number of instances pseudo-labeled as class $c$, respectively.

To avoid a trivial solution in which all the instance embeddings are assigned to a single prototype, an equipartition constraint is added to ensure that instances are equally distributed among prototypes within a class. By adding the entropy regularization term~\cite{cuturi2013sinkhorn} with a parameter $\kappa$ that controls the smoothness of assignment, the objective for obtaining the assignment matrix for class $c$ is as follows:
\begin{equation}
\label{prototype_assignment_obj}
\max_{{\mathbf{L}}_{c}}
    \text{Tr}\left( 
    {\mathbf{L}}_{c}^{\top} \mathbf{Q}_c 
    \right) 
    +\kappa~\mathcal{H}(\mathbf{L}_{c}),
    \quad \textit{s.t.} \quad
    \mathbf{L}_{c} \cdot \mathbf{1}^{n_c}=\frac{n_c}{K} \cdot \mathbf{1}^K,
\end{equation} which turns into an optimal transport problem. The solution can be computed by a few iterations of the \textit{Sinkhorn-Knopp} algorithm~\cite{cuturi2013sinkhorn}, which outputs the re-normalization vectors $\mathbf{u}\in \mathbb{R}^{K}$ and $\mathbf{v} \in \mathbb{R}^{n_c}$:
\begin{equation}\label{sinkhorn_knopp}
    \mathbf{L}_{c} = 
    \operatorname{diag}(\mathbf{u})
    \exp\left({\frac{\mathbf{Q}_{c}}{\kappa}}\right)
    \operatorname{diag}(\mathbf{v}).
\end{equation}

After obtaining the soft assignment matrix, the network is trained so that similarities between prototypes and instance embeddings correspond to the soft assignment matrix. The prototypes and instance embeddings are simultaneously optimized by minimizing the following cross-entropy loss between the similarity and assignment matrix:
\begin{equation}
\label{instance_loss}
    \mathcal{L}_{inst} = -\frac{1}{N \cdot K}
    \sum_{i=1}^{N} 
    \sum_{j=1}^{K}
    \mathbf{L}^{i,j}_{\hat{c}_i} \cdot \log \frac{
        \exp{
        \left(\mathbf{z}_i \cdot \mathbf{p}_{\hat{c}_i, j}/ \tau
        \right)
        }
    }
    {\sum_{c}^{C}\sum_{k}^{K}
    \exp{\left(
    \mathbf{z}_i \cdot \mathbf{p}_{c, k}/ \tau
    \right)}},
\end{equation} where $\hat{c}_i$ is the pseudo-label for the $i^{th}$ instance, and $\mathbf{L}^{i,j}_{\hat{c}_i}$ is a soft assignment of the instance to the $j^{th}$ prototype of the pseudo-labeled class. Also, $\tau$ is a temperature parameter, and $C$ denotes the number of classes, including the background class. For each instance $\mathbf{z}_i$, minimizing $\mathcal{L}_{inst}$ increases its similarity with assigned prototypes $\mathbf{p}_{\hat{c}_i, j}$, and decreases its similarity with all the others. Note that the loss is computed on both the source and target domain features using the same prototypes to reduce the domain gap.

As a result, instance embeddings are grouped around their assigned prototypes. This produces corruption-resistant instance features by promoting instance embeddings with similar semantics and variable weather corruption to become closer. The final objective of our method is as follows:
\begin{equation}
\mathcal{L} = \mathcal{L}_{sup} + \alpha \mathcal{L}_{img} + \beta \mathcal{L}_{inst}.
\end{equation}

%% file: _IV.Experiments/4.0.Overview.tex
In this section, we validate that our unsupervised domain adaptation method can effectively enhance object detection performance in real-world environments with adverse weather. Using publicly available datasets and our own datasets as evaluation, the results of our method are quantitatively and qualitatively compared to those of other methods for object detection under adverse weather conditions. In addition, ablation studies are conducted to assess the validity of each component of our methodology.

%% file: _IV.Experiments/4.1.Datasets.tex
\begin{table*}[t]
\centering
\renewcommand{\arraystretch}{1.2}
\caption{
 Quantitative results on both synthetic and real-world datasets. mAP (\%) is used as an evaluation metric. \textbf{Synthetic Data} column indicates whether synthetically generated adverse weather data are utilized during training, and the \textbf{Target Data} column indicates whether unlabeled data from the target domain is incorporated during training. Our method outperforms other methods when applied to datasets with real-world adverse weather conditions.
}
\label{tab:result}
\resizebox{1.0\textwidth}{!}{%
    \begin{tabular}{cccccccccc}
    \toprule
    \multirow{4}{*}{\textbf{Method}} & \multirow{4}{*}{\textbf{Source Data}} & \multirow{4}{*}{\textbf{Synthetic Data}} & 
    \multirow{4}{*}{\textbf{Target Data}} & \multicolumn{4}{c}{\textbf{Rainy}} & \multicolumn{2}{c}{\textbf{Snowy}} \\
    \cmidrule(rl){5-8}\cmidrule(rl){9-10}
    & & & & \multicolumn{2}{c}{\textbf{Synthetic}} & \multicolumn{2}{c}{\textbf{Real-World}} & \multicolumn{2}{c}{\textbf{Real-World}} \\
    \cmidrule(rl){5-6}\cmidrule(rl){7-8}\cmidrule(rl){9-10}
    & & & & \textit{RainCityscapes}~\cite{hu2019depth} & \textit{Rain Rendering}~\cite{halder2019physics} & \textit{BDD 100K}~\cite{yu2020bdd100k} & \textit{Our Dataset} &\textit{BDD 100K}~\cite{yu2020bdd100k} & \textit{Our Dataset}\\
    \midrule \midrule
    \textit{Source Only} & \cmark & \xmark & \xmark & 35.0 & 31.4 & 31.6 & 49.4 & 27.9 & 57.8 \\
    \textit{Physics-based}~\cite{halder2019physics} & \cmark & \cmark & \xmark & \textbf{40.5} & 41.8 & 22.1 & 35.1 & 18.1 & 42.0 \\
    \textit{MPRNet}~\cite{zamir2021multi} & \cmark & \cmark & \xmark & 37.7 & \textbf{46.9} & 12.8 & 38.6 & 12.3 & 41.8 \\
    \textit{SADA}~\cite{chen2021scale} & \cmark & \xmark & \cmark & 38.7 & 40.1 & 29.1 & 48.2 & 27.6 & 53.5 \\
    \textit{SWDA}~\cite{saito2019strong} & \cmark & \xmark & \cmark & 37.7 & 36.7 & 31.1 & 49.3 & 28.4 & 58.4 \\
    \textit{Ours} & \cmark & \xmark& \cmark & 37.7 & 35.6 & \textbf{34.5} & \textbf{51.2} & \textbf{30.3} & \textbf{62.6}  \\
    \bottomrule
    \end{tabular}
    }
\end{table*}

\begin{figure*}[t]
\begin{center}
\includegraphics[width=1.0\linewidth]{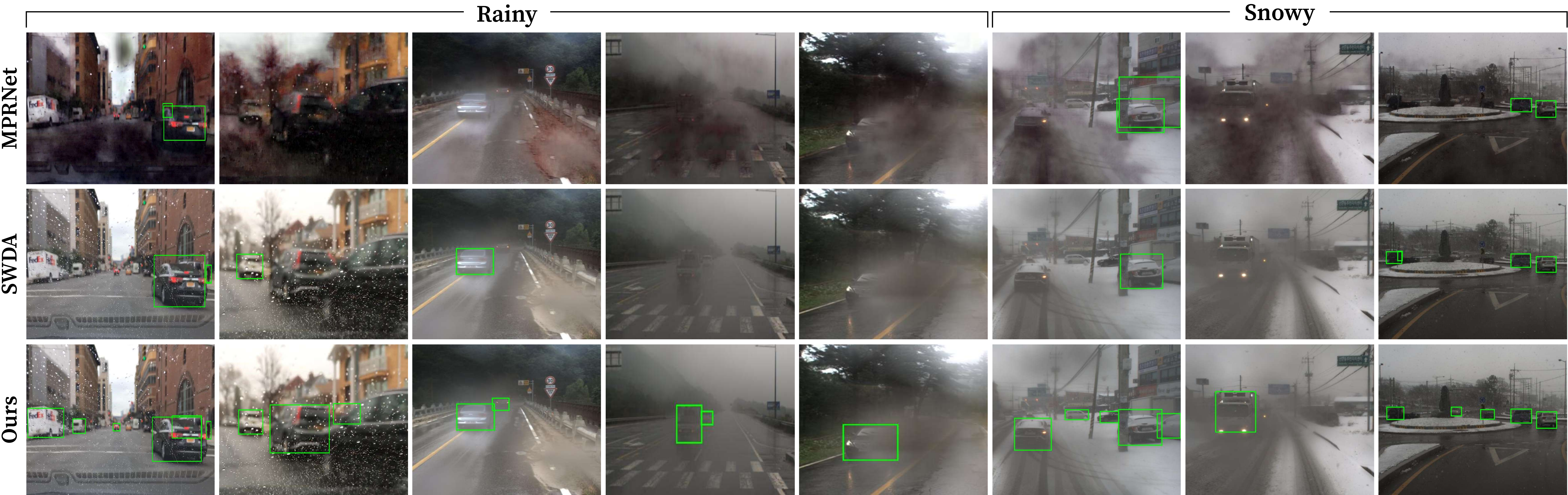}
\end{center}
\caption{
Qualitative results on real-world target datasets. Compared to other methods, \textit{Ours} successfully detects the objects even in the presence of severe weather corruption and style variations. Even though \textit{MPRNet} removes raindrops in the first two images, the detector performance remains low, indicating images generated by the removal network do not consistently help object detection. In the remaining images, the removal network fails to remove corruptions and instead creates some artifacts due to the disparity between real weather data and synthetic weather data, which \textit{MPRNet} was trained on. While \textit{SWDA} directly adapts the network to the target domain, it fails to detect objects under severe weather corruption and environmental differences. More qualitative results are available in our multimedia material. 
}
\label{fig:result}
\vspace{-0.1in}
\end{figure*}

Our source domain dataset is \textit{Cityscapes}~\cite{cordts2016cityscapes}, which consists of real-world urban driving images captured under clear weather conditions. For the target domain dataset, multiple datasets are used to validate the efficacy of our method in various environments with adverse weather conditions. Using two synthetic weather datasets, we investigate the efficacy of other methods employing synthetic data or UDA in synthetic weather contexts. On clear images of the \textit{Cityscapes}, the \textit{Rain Rendering}~\cite{halder2019physics} generates synthetic rain images with a physical particle simulator for each, and \textit{RainCityscapes}~\cite{hu2019depth} generates rain and fog effects subject to scene depth. To validate the efficacy of methods in real-world environments, \textit{BDD 100K}~\cite{yu2020bdd100k}, a real-world driving dataset captured in various weather conditions, is employed. The rainy and snowy subsets are used as our target domain to evaluate the efficacy of methods under diverse weather conditions.

To further validate our model across a wider range of environments, we collect \textit{Our Dataset} in adverse weather conditions using our platform, which is equipped with an external vehicle RGB camera~\cite{taodac,seo2023learning}. Several datasets with adverse weather scenarios primarily focus on urban scenes and lack diversity in background environments~\cite{kenk2020dawn, pitropov2021canadian}. To address significant domain gaps, our datasets include data collected from rural and mountainous environments. In comparison to the \textit{BDD 100K}, which uses a camera mounted inside the windshield of the vehicle, our dataset utilizes an external camera that is consistent with the source domain dataset. In addition, raindrops and snowfalls on the lens result in much more severe blurring of the images. 
Our dataset is divided into \textit{Rainy} and \textit{Snowy} subsets. The rainy subset consists of 2845 training images and 677 validation images, and the snowy subset consists of 1656 training images and 598 validation images. Our dataset comprises three classes: \textit{person}, \textit{car}, and \textit{motorcycle}. For alignment with \textit{Cityscapes} dataset, \textit{Cityscapes} classes are assigned to our dataset during the experiment as follows: 1) \textit{person}, \textit{rider} to \textit{person}, 2) \textit{car}, \textit{truck}, \textit{bus} to \textit{car}, and 3) \textit{motorcycle}, \textit{bicycle} to \textit{motorcycle}.

%% file: _IV.Experiments/4.2.Experimental_setup.tex
\noindent\textbf{Implementation Details.}
We adopt Faster R-CNN with ImageNet-pretrained ResNet-50~\cite{he2016deep} and FPN as our object detection network. Initially, the model is trained with only the $\mathcal{L}_{\text{sup}}$ using source data in order to obtain pseudo-labels, and $\mathcal{L}_{\text{img}}$ and $\mathcal{L}_{\text{inst}}$ are applied after 7.5k iterations with ${\alpha=1.0}$, ${\beta=1.0}$. The image-level style alignment is performed on the FPN features at levels $P_4$ and $P_5$, with ${\lambda}$ of GRL set as $0.01$. The instance-level weather alignment is performed on features at levels $P_2$ and $P_3$. We execute three iterations of the Sinkhorn-Knopp algorithm with smoothness parameter ${\kappa=0.05}$, each taking about 0.1 milliseconds. The number of prototypes $K$ for each class is set to $5$, with other hyperparameters empirically set to ${\tau=0.05}$, ${\delta=0.8}$, ${D=128}$.

During training, stochastic gradient descent~(SGD) is used as an optimizer with a weight decay of $5e^{-4}$ and momentum of $0.9$. Each batch consists of 16 images, eight from the source domain and eight from the target domain. We resized all the input images so that the shorter side has a length of 800 pixels and applied random horizontal flipping with a probability of 0.5. The entire model is trained with an initial learning rate of $2.5e^{-3}$ for $9.5k$ iterations and then reduced to $2.5e^{-4}$ for another $2k$ iterations.

\noindent\textbf{Comparison Methods.}
To demonstrate the efficacy of our method under real-world adverse weather conditions, our method is compared to other detection methods designed for such conditions. For comparison with the method utilizing synthetic weather data, \textit{Physics-based}~\cite{halder2019physics} is adopted, which trains the model in a supervised manner using synthetic rainy images. Note that \textit{Physics-based} and \textit{Rain Rendering} dataset uses the same method to synthesize rain. The rain removal network, \textit{MPRNet}~\cite{zamir2021multi}, is also utilized. The removal network is trained using synthetic paired images of \textit{Cityscapes} and \textit{Rain Rendering}, and evaluation is conducted on restored images from the network using a detector trained only with clear source domain data. We also include results from several UDA methods that were trained from scratch. \textit{SADA}~\cite{chen2021scale} directly aligns source and target features at both the image-level and instance-level through adversarial training, whereas \textit{SWDA}~\cite{saito2019strong} focuses solely on aligning image-level features. However, none of the aforementioned methods address the style and weather gaps separately.

\noindent\textbf{Evaluation Metric.}
The mean Average Precision~(mAP) of all categories is used for evaluation with an Intersection over Union~(IoU) threshold of $0.5$ to compute the Average Precision~(AP). Due to class imbalance issues in our dataset, class-agnostic AP is calculated for all the bounding boxes with an IoU threshold of $0.5$ for a fair comparison.

%% file: _IV.Experiments/4.3.Experimental_Results.tex
\begin{table}[thb]
\centering
\renewcommand{\arraystretch}{1.2}
\caption{
 Results of the ablation studies. mAP (\%) is used as an evaluation metric. Incorporating each module improves detection performance.
}
\label{tab:ablation}
\large{
\resizebox{1.0\linewidth}{!}{%
    \begin{tabular}{cccccc}
        \toprule
        \multicolumn{2}{c}{\textbf{Module}} & \multicolumn{2}{c}{\textbf{Rainy}} & \multicolumn{2}{c}{\textbf{Snowy}} \\
        \cmidrule(lr){1-2} \cmidrule(lr){3-4} \cmidrule(lr){5-6}
        \textbf{Style} & \multicolumn{1}{c}{\textbf{Weather}} & \textit{BDD 100K}~\cite{yu2020bdd100k} & \textit{Our Dataset} & \textit{BDD 100K}~\cite{yu2020bdd100k} & \textit{Our Dataset} \\
        \midrule 
        \xmark & \xmark & 31.6 & 49.4 & 27.9 & 57.8\\
        \multicolumn{1}{l}{\cmark \hspace{5pt} \textbf{(w.o. CBAM)}} & \xmark & 31.5 & 50.2 & 27.1 & 59.2\\
        \multicolumn{1}{l}{\cmark \hspace{11pt} \textbf{(w. CBAM)}} & \xmark & 32.4 & 50.7 & 28.7 & 59.6\\
        \xmark & \cmark & 33.7 & 51.0 & 29.3 & 61.5\\
        \multicolumn{1}{l}{\cmark \hspace{11pt} \textbf{(w. CBAM)}} & \cmark & \textbf{34.5} & \textbf{51.2} & \textbf{30.3} & \textbf{62.6}\\
        \bottomrule
    \end{tabular}
    }
}
\vspace{-0.1in}
\end{table}

\noindent\textbf{Comparisons with Other Methods.}
The quantitative and qualitative results are summarized in Table~\ref{tab:result} and Fig~\ref{fig:result}. In both rainy and snowy conditions, our approach outperforms other methods when applied to real-world datasets. Existing UDA methods such as \textit{SADA} and \textit{SWDA} show a significant improvement in performance on synthetic datasets, but their performance on real-world datasets is only marginally improved or even decreases. This implies that existing methods that globally align distributions are ineffective when adapting to real-world datasets with a large style gap and severe weather corruption, in contrast to synthetic datasets with a small domain gap from synthetic weather.

In addition, \textit{SWDA} outperforms \textit{SADA}, despite the fact that \textit{SADA} incorporates instance-level alignment while \textit{SWDA} focuses solely on image-level alignment. This suggests that directly aligning the instance-level features that are severely contaminated in real adverse weather conditions reduces performance, necessitating the use of alternative alignment methods. Using image-level style alignment and instance-level weather alignment, our method optimizes feature alignment by resolving both style and weather gaps, thereby improving detection performance.

\noindent\textbf{Efficacy of Synthetic Weather Dataset.}
Methods that utilize synthetic weather images during training perform well when evaluated on synthetic data. However, their performance decreases when evaluated on real-world data, indicating that synthetic weather fails to accurately represent the complexities of real weather. The use of removal networks on real-world datasets also has a negative effect on performance, despite requiring more computation. As shown in Fig~\ref{fig:result}, the removal network trained on synthetic data has difficulty restoring a clear image from real-world images under both rainy and snowy conditions, showing its inability to remove complex real rain and generalize to other weather conditions. In addition, detection performance decreases even in visually restored areas. This suggests that the features obtained through the removal network do not contribute to improving the detection performance. By directly adapting to downstream tasks, our method achieves superior performance on real-world datasets without relying on synthetic priors.

%% file: _IV.Experiments/4.4.Analysis.tex
\setlength{\aboverulesep}{0pt}
\setlength{\belowrulesep}{0pt}

\noindent\textbf{Ablation Studies on Each Component.}
To validate the efficacy of each component, we conducted an ablation analysis on real-world datasets. The results are shown in Table~\ref{tab:ablation}. Image-level style alignment increases performance, indicating that high-level feature alignment bridges the domain gap between image-level features effectively. Particularly, there is a significant performance increase when CBAM is present, implying that CBAM improves alignment by focusing on essential features.
Incorporating instance-level weather alignment further enhances performance. This suggests that instance features obtained by self-supervised contrastive learning are robust to weather corruption and domain-invariant. Overall, combining both components achieves the highest performance, which effectively addresses both gaps.
\begin{figure}[t]
\begin{center}
\includegraphics[width=1.0\linewidth]{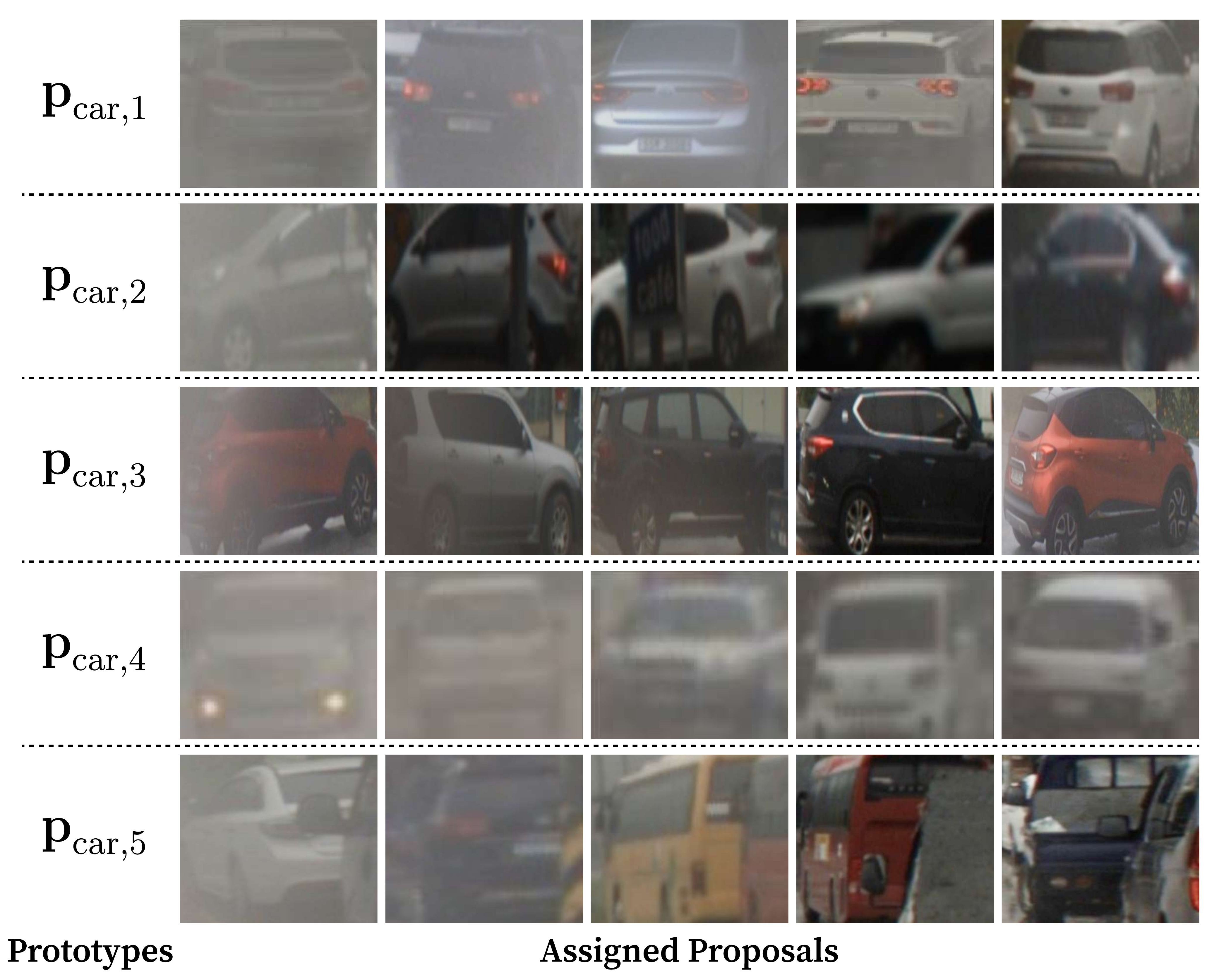}
\end{center}
\caption{
Visualization of proposals assigned to each prototype in our rainy dataset. Each row displays the proposals whose instance embedding is highly similar to each car class prototype. Similar-shaped objects with diverse corruption and styles are assigned to identical prototypes, indicating the effectiveness of prototype-based contrastive learning. For example, the first row contains car proposals captured from a rear-view perspective and showing varying degrees of corruption.
}
\vspace{-0.1in}
\label{fig:instance}
\end{figure}

\noindent\textbf{Efficacy of Instance-level Weather Alignment.}
To evaluate the impact of the instance-level weather alignment on feature embeddings, we visualize proposals whose instance embeddings are highly similar to each of the car class prototypes. As shown in Fig~\ref{fig:instance}, proposals with similar object shapes but varying degrees of corruption are assigned to the same prototype. This demonstrates that our weather align module contributed to extracting semantically meaningful features that are resilient to corruption. Moreover, the fact that objects with similar shapes are gathered together in the same prototype demonstrates the efficacy of employing multiple prototypes to address intra-class variation.

%% file: _V.Conclusion/5.1.Conclusion.tex
This paper presents a novel framework for domain adaptive object detection that improves robustness under real-world adverse weather conditions. The proposed method effectively addresses two distinct aspects of the domain gap, the style gap and the weather gap, by using image-level style alignment and instance-level weather alignment, respectively. Diverging from previous approaches that were mostly evaluated in synthetic datasets, our method shows robust performance on real-world datasets, which have been validated through extensive experiments. 

We believe that our method can expand the range of applications for machines as it can detect objects in a variety of real adverse weather conditions. To make our method more applicable to real-world applications, we are investigating techniques that can adapt to dynamic adverse weather conditions during inference time. Furthermore, we also aim to expand our framework with minimal requirements for target domain data.
